\documentstyle{article}
\makeatletter
\relax
\@writefile{toc}{\string\contentsline\space {section}{
\string\numberline\space {1}Introduction}{1}}
\@writefile{toc}{\string\contentsline\space {section}{
\string\numberline\space {2}Distance Vectors}{1}}
\@writefile{toc}{\string\contentsline\space {section}{
\string\numberline\space {3}Co-occurrence Vectors}{2}}
\@writefile{toc}{\string\contentsline\space {section}{
\string\numberline\space {4}Experimental Results}{3}}
\@writefile{toc}{\string\contentsline\space {subsection}{
\string\numberline\space {4.1}Word Sense Disambiguation}{3}}
\@writefile{toc}{\string\contentsline\space {subsection}{
\string\numberline\space {4.2}Learning of {\string\pit\space {
\string\pit\space positive}-or-{\string\pit\space negative}}}{3}}
\@writefile{toc}{\string\contentsline\space {subsection}{
\string\numberline\space {4.3}Supplementary Data}{5}}
\@writefile{toc}{\string\contentsline\space {section}{
\string\numberline\space {5}Conclusion}{5}}
\def\xcomment#1#2{#1}
\newcount\myfignmb \myfignmb=1
\def\newfignmb#1{\newcount#1 #1=\the\myfignmb \advance\myfignmb by 1}
\newcount\mytablenmb \mytablenmb=1
\def\newtablenmb#1{\newcount#1 #1=\the\mytablenmb \advance\mytablenmb by 1}

\def\Comment#1{}
\def\setbs{\baselineskip=0.1667in}
\newlength\titlebox
\let\maketitle\relax
\let\@maketitle\relax
\def\maketitle{\par
  \begingroup
    \def\thefootnote{\fnsymbol{footnote}}
    \def\@makefnmark{\hbox to 0pt{$^{\@thefnmark}$\hss}}
    \twocolumn[\@maketitle]%
    \@thanks%
  \endgroup
  \setcounter{footnote}{0}%
  \let\maketitle\relax%
  \let\@maketitle\relax%
  \gdef\@thanks{}\gdef\@author{}\gdef\@title{}%
  \let\thanks\relax%
}

\def\toappear{}
\def\@maketitle{%
\hbox to \textwidth{\unitlength=1mm\begin{picture}(0,0)(0,0)
\put(0,5){\makebox(0,0)[lb]{\normalsize\tt\toappear}}\end{picture}\hfil}
  \vbox to \titlebox{%
    \hsize\textwidth
    \linewidth\hsize
    \vskip 4.5mm
    \centering
    {\LARGE\bf \@title \par} 
    \vskip 3mm
    {\def\and{\unskip\enspace{\rm and}\enspace}
     \def\And{\end{tabular}\hss\egroup\hskip 10mm
              \hbox to 0pt\bgroup\hss \begin{tabular}[t]{c}\large\bf}
     \def\AND{\end{tabular}\hss\egroup\hfil\hfil\egroup
              \vskip 3mm
              \hbox to \linewidth\bgroup\large\hfil\hfil
              \hbox to 0pt\bgroup\hss\begin{tabular}[t]{c}\large\bf}
     \hbox to \linewidth\bgroup\large\hfil\hfil
     \hbox to 0pt\bgroup\hss\begin{tabular}[t]{c}\large\bf\@author
                            \end{tabular}\hss\egroup
     \hfil\hfil\egroup}
    \vskip 8.5mm
    \vfil
}}

\def\@citex[#1]#2{\if@filesw\immediate\write\@auxout{\string\citation{#2}}\fi
  \def\@citea{}\@cite{\@for\@citeb:=#2\do
    {\@citea\def\@citea{;\penalty\@m\ }\@ifundefined
       {b@\@citeb}{{\bf ?}\@warning
       {Citation `\@citeb' on page \thepage \space undefined}}%
{\csname b@\@citeb\endcsname}}}{#1}}
\let\@internalcite\cite
\def\cite{\def\citename##1{##1 }\@internalcite}
\def\shortcite{\def\citename##1{}\@internalcite}
\def\newcite{\leavevmode\def\citename##1{{##1} (}\@internalciteb}
\def\@citexb[#1]#2{\if@filesw\immediate\write\@auxout{\string\citation{#2}}\fi
  \def\@citea{}\@newcite{\@for\@citeb:=#2\do
    {\@citea\def\@citea{;\penalty\@m\ }\@ifundefined
       {b@\@citeb}{{\bf ?}\@warning
       {Citation `\@citeb' on page \thepage \space undefined}}%
\hbox{\csname b@\@citeb\endcsname}}}{#1}}
\def\@internalciteb{\@ifnextchar [{\@tempswatrue\@citexb}{%
\@tempswafalse\@citexb[]}}
\def\@newcite#1#2{{#1\if@tempswa, #2\fi)}}
\def\@biblabel#1{\def\citename##1{##1}[#1]\hfill}
\def\@cite#1#2{({#1\if@tempswa , #2\fi})}

\def\Comment#1{}
\def\thebibliography#1{\vskip\parskip%
\vskip\baselineskip%
\def\baselinestretch{1}%
\ifx\@currsize\normalsize\@normalsize\else\@currsize\fi%
\vskip-\parskip%
\vskip-\baselineskip%
\section*{References\@mkboth
 {References}{References}}\list
 {}{\setlength{\labelwidth}{0pt}\setlength{\leftmargin}{\parindent}
 \setlength{\itemindent}{-\parindent}}
 \def\newblock{\hskip .11em plus .33em minus -.07em}
 \sloppy\clubpenalty4000\widowpenalty4000
\baselineskip=0.16in
\parskip=0mm
 \sfcode`\.=1000\relax}

\def\thesourcebibliography#1{\vskip\parskip%
\vskip\baselineskip%
\def\baselinestretch{1}%
\ifx\@currsize\normalsize\@normalsize\else\@currsize\fi%
\vskip-\parskip%
\vskip-\baselineskip%
\section*{Sources of Attested Examples\@mkboth
 {Sources of Attested Examples}{Sources of Attested Examples}}\list
 {}{\setlength{\labelwidth}{0pt}\setlength{\leftmargin}{\parindent}
 \setlength{\itemindent}{-\parindent}}
 \def\newblock{\hskip .11em plus .33em minus -.07em}
 \sloppy\clubpenalty4000\widowpenalty4000
 \sfcode`\.=1000\relax}

\def\@lbibitem[#1]#2{\item[]\if@filesw
      { \def\protect##1{\string ##1\space}\immediate
        \write\@auxout{\string\bibcite{#2}{#1}}\fi\ignorespaces}}
\def\@bibitem#1{\item\if@filesw \immediate\write\@auxout
       {\string\bibcite{#1}{\the\c@enumi}}\fi\ignorespaces}

\def\tmpca{$\circ$}
\def\tmpcb{}
\def\tmppos{{\it positive}}
\def\tmpneg{{\it negative}}
\def\tmpd{$\bullet$}
\def\cedsizehead{60K}
\def\cedsizedef{1.6M}
\def\tmpul{0.5mm}

\def\graphnezero#1{{\unitlength=\tmpul\small
  \begin{picture}(100,108)(0,-5)#1\end{picture}}}

\def\graphvd#1{
  \graphnezero{
    \put(0,0){\line(1,0){100}}
    \put(0,0){\line(0,1){100}}
    \multiput(0,10)(0,10){10}{\line(1,0){2}}
    \put(-2,50){\makebox(0,0)[r]{\small 50\%}}
    \put(-2,100){\makebox(0,0)[r]{\small 100\%}}
    {\small\def\tmpya{-2}
     \multiput(10,0)(20,0){5}{\line(0,1){2}}
     \multiput(20,0)(40,0){3}{\line(0,1){3}}
     \multiput(40,0)(40,0){2}{\line(0,1){2}}
     \put(20,\tmpya){\makebox(0,0)[t]{10}}
     \put(60,\tmpya){\makebox(0,0)[t]{100}}
     \put(100,\tmpya){\makebox(0,0)[t]{1000}}}
    #1}}
\def\graphcs#1{{\unitlength=\tmpul\small
  \begin{picture}(106.7,110)(0,-7)
    \put(0,0){\line(1,0){106.7}}
    \put(0,0){\line(0,1){100}}
    \multiput(0,10)(0,10){10}{\line(1,0){2}}
    {\small \put(-2,50){\makebox(0,0)[r]{50\%}}
            \put(-2,100){\makebox(0,0)[r]{100\%}}}
    \multiput(20,0)(20,0){5}{\line(0,1){2}}
    \multiput(6.7,0)(20,0){6}{\line(0,1){1}}
    \multiput(13.3,0)(20,0){5}{\line(0,1){1}}
    {\small\def\tmpya{-2}
     \put(20,\tmpya){\makebox(0,0)[t]{$10^3$}}
     \put(40,\tmpya){\makebox(0,0)[t]{$10^4$}}
     \put(60,\tmpya){\makebox(0,0)[t]{$10^5$}}
     \put(80,-3.5){\makebox(0,0)[t]{\bf 1M}}
     \put(100,-3.5){\makebox(0,0)[t]{10M}}}
    #1
  \end{picture}}}
\def\tcomment#1{#1}
\def\comment#1{#1}
\def\gcomment#1{}
\def\Fbib#1{}

\makeatother
\begin{document}
\setcounter{page}{304}
\twocolumn\normalsize
\setbs
\def\tcomment#1{}
\def\comment#1{#1}
\def\toappear{}
\title{{\large\bf
CO\raise0.7mm\hbox{-}OCCURRENCE VECTORS FROM CORPORA VS.\\[-2mm]
DISTANCE VECTORS FROM DICTIONARIES\vspace{2mm}
}}
\author{
Yoshiki Niwa \and Yoshihiko Nitta\\[2mm]
Advanced Research Laboratory, Hitachi, Ltd.\\
Hatoyama, Saitama \verb@350-03@, Japan\\
\verb+\{niwa2, nitta\}@harl.hitachi.co.jp+
}

\maketitle
\subsection*{Abstract}
A comparison was made of vectors derived by using ordinary co-occurrence
statistics from large text corpora and
of vectors derived by measuring the inter-word distances in dictionary
definitions.
The precision of word sense disambiguation by
using co-occurrence vectors from the 1987 Wall Street Journal
(20M total words) was higher than that by using
distance vectors from the Collins English Dictionary
({\cedsizehead} head words + {\cedsizedef} definition words).
However, other experimental results suggest that distance vectors
contain some different semantic information from co-occurrence vectors.

\setcounter{section}{0}\def\tcomment#1{}

\section{Introduction}

Word vectors reflecting word meanings are expected to enable
numerical approaches to semantics.
Some early attempts at vector representation in psycholinguistics
were the {\it semantic differential} approach \cite{Osgood+al:1957}
and the {\it associative distribution} approach \cite{Deese:PsyRev62}.
However, they were derived manually through psychological experiments.
An early attempt at automation was made by Wilks {\it et al.}
\shortcite{Wilks+al:MT90} using co-occurrence statistics.
Since then, there have been some promising results from using
co-occurrence vectors, such as word sense disambiguation
\cite{Schuetze:NIPS93}, and word clustering \cite{Pereira+al:ACL93}.

However, using the co-occurrence statistics requires a huge corpus
that covers even most rare words.
We recently developed word vectors that are derived from an ordinary
dictionary by measuring the inter-word distances in the word definitions
\cite{Niwa+Nitta:MCCS93}.
This method, by its nature, has no problem handling rare words.
In this paper we examine the usefulness of these {\it distance vectors}
as semantic representations by comparing them with co-occurrence vectors.

\section{Distance Vectors}

\newfignmb{\figrn}
A reference network of the words in a dictionary (Fig. \the\figrn)
is used to measure the distance between words.
The network is a graph that shows which words are used in
the definition of each word \cite{Nitta:LNL88}.
The network shown in Fig.\,{\the\figrn} is for a very small portion
of the reference network for the Collins English Dictionary
(1979 edition) in the CD-ROM I \cite{ACL-CD-ROM-1},
with {\cedsizehead} head words + {\cedsizedef} definition words.
\vskip3mm
\centerline{
\normalsize\unitlength=1mm
\begin{picture}(80,36)(-40,-18)
\put(-4,5){word}\put(-15,14){writing}\put(6,14){unit}
\put(5,-8){dictionary}
\put(20,0){alphabetical}\put(20,-17){book}
\put(-5,-7){\line(1,0){9}}
\put(-8,-6){\line(2,3){7}}
\put(7,-5){\line(-2,3){6}}
\put(-2,8){\line(-1,1){5}}
\put(2,8){\line(1,1){5}}
\put(-16,-6){\line(-1,1){5}}
\put(16,-6){\line(1,1){5}}
\put(-16,-9){\line(-1,-1){5}}
\put(16,-9){\line(1,-1){5}}
\put(-19,-8){language}
\put(-40,0){communication}
\put(-25,-17){people}
\put(13,14){\((O_1)\)}
\put(28,-16){\((O_2)\)}
\put(-32.5,-16){\((O_3)\)}
\end{picture}}
\vskip2mm
\centerline{{\bf Fig. \the\figrn}\ \ Portion of a reference network.}
\vskip3mm

For example, the definition for {\it dictionary} is ``a book in which the
words of a language are listed alphabetically ...\hspace{2mm}.''
The word {\it dictionary} is thus linked to the words
{\it book, word, language}, and {\it alphabetical}.

A word vector is defined as the list of distances from a word to a
certain set of selected words, which we call {\it origins}.
The words in Fig.\,{\the\figrn} marked with $O_i$ ({\it unit, book},
and {\it people}) are assumed to be origin words.
In principle, origin words can be freely chosen.
In our experiments we used middle frequency words:
the 51st to 1050th most frequent words
in the reference Collins English Dictionary (CED).

The distance vector for {\it dictionary} is derived as follows:
\[\hbox{\it dictionary}\ \Rightarrow\
  \left(\begin{array}{c} 2\\[1mm]1\\[1mm]2\end{array}\right)
  \begin{array}{c}
     \hbox{\(\cdot\cdot\cdot\) distance ({\it dict.}, \(O_1\))}\\[1mm]
     \hbox{\(\cdot\cdot\cdot\) distance ({\it dict.}, \(O_2\))}\\[1mm]
     \hbox{\(\cdot\cdot\cdot\) distance ({\it dict.}, \(O_3\))}
  \end{array}\hspace{3mm}.\]
The $i$-th element is the
distance (the length of the shortest path) between {\it dictionary} and
the $i$-th origin, $O_i$.
To begin, we assume every link has a constant length of 1.
The actual definition for link length will be given later.

If word A is used in the definition of word B, these words are expected
to be strongly related.
This is the basis of our hypothesis that the distances in the reference
network reflect the associative distances between words
\cite{Nitta:Korea93}.

{\bf Use of Reference Networks}\hspace{3mm}
Reference networks have been successfully used as
neural networks (by V\'{e}ronis and Ide \shortcite{Veronis+Ide:COLING90}
for word sense disambiguation)
and as fields for artificial association, such as spreading activation
(by Kojima and Furugori \shortcite{Kojima+Furugori:EACL93} for
context-coherence measurement).
The distance vector of a word can be considered to be a list of the
activation strengths at the origin nodes when the word node is activated.
Therefore, distance vectors can be expected to convey almost the same
information as the entire network, and clearly they are much easier
to handle.

{\bf Dependence on Dictionaries}\hspace{3mm}
As a semantic representation of words, distance vectors are expected to
depend very weakly on the particular source dictionary.
We compared two sets of distance vectors,
one from {\small LDOCE} \cite{LDOCE} and the other from
{\small COBUILD} \cite{COBUILD}, and verified
that their difference is at least smaller than the difference of
the word definitions themselves \cite{Niwa+Nitta:MCCS93}.

We will now describe some technical details about the derivation
of distance vectors.

{\bf Link Length}\hspace{3mm}
Distance measurement in a reference network depends on the
definition of link length.
Previously, we assumed for simplicity that every link
has a constant length.
However, this simple definition seems unnatural because it does not
reflect word frequency.
Because a path through low-frequency words (rare words) implies
a strong relation, it should be measured as a shorter path.
Therefore, we use the following definition of link length,
which takes account of word frequency.

\[ \hbox{length}\ ( \hbox{W}_1,\hbox{W}_2 )
   \parbox[t]{9mm}{\centerline{\(=\)}\vskip-1mm\centerline{\small\it def}}
   - \log
   \left(\frac{\hbox{n}^2}{\hbox{N}_1 \cdot \hbox{N}_2}\right)
\]

\newfignmb{\figlink}
This shows the length of the links between words W\(_i (i=1,2)\)
in Fig.\,{\the\figlink},
where N\(_i\) denotes the total number of links from and to W\(_i\)
and n denotes the number of direct links between these two words.
\vskip3mm
\vbox{
\centerline{\large\unitlength=0.7mm
\begin{picture}(80,40)(0,0)
\put(20,20){\circle{10}}
\put(60,20){\circle{10}}
\put(16,18){\(\hbox{W}_1\)}
\put(56,18){\(\hbox{W}_2\)}
\put(25,20){\line(1,0){13}}\put(55,20){\line(-1,0){13}}
\put(38,18){\makebox(4,4){n}}
\put(30,23){\line(1,0){20}}
\put(30,23){\line(-5,-2){5.1}}\put(50,23){\line(5,-2){5.1}}
\put(30,17){\line(1,0){20}}
\put(30,17){\line(-5,2){5.1}}\put(50,17){\line(5,2){5.1}}
\put(13,35){\(\hbox{N}_1\)}
\put(65,35){\(\hbox{N}_2\)}
\put(23.5,23.5){\line(1,1){8}}
\put(20,25){\line(0,1){15}}
\put(16.5,23.5){\line(-1,1){11}}
\put(15,20){\line(-1,0){15}}
\put(16.5,16.5){\line(-1,-1){11}}
\put(20,15){\line(0,-1){15}}
\put(23.5,16.5){\line(1,-1){8}}
\put(56.5,23.5){\line(-1,1){8}}
\put(60,25){\line(0,1){15}}
\put(63.5,23.5){\line(1,1){11}}
\put(65,20){\line(1,0){15}}
\put(63.5,16.5){\line(1,-1){11}}
\put(60,15){\line(0,-1){15}}
\put(56.5,16.5){\line(-1,-1){8}}
\end{picture}}
\vskip2mm
\centerline{{\bf Fig. \the\figlink}\ \ Links between two words.}
}
\vskip3mm

{\bf Normalization}\hspace{3mm}
Distance vectors are normalized by first
changing each coordinate into its deviation in the coordinate:
\[v = (v_i)\hspace{5mm}\rightarrow\hspace{5mm}
  v' = \left( \frac{v_i - a_i}{\sigma_i} \right)\;, \]
where \(a_i\) and \(\sigma_i\) are the average and the standard deviation
of the distances from the \(i\)-th origin.
Next, each coordinate is changed into its deviation in the vector:
\[v' = (v'_i)\hspace{5mm}\rightarrow\hspace{5mm}
  \overline{v} = \left( \frac{v'_i - \overline{v'}}{\sigma'}\right)\;,\]
where \(\overline{v'}\) and \(\sigma'\) are the average and the standard
deviation of \(v'_i\; \hbox{\small\((i=1,...)\)}\).

\section{Co-occurrence Vectors}

We use ordinary co-occurrence statistics and
measure the co-occurrence likelihood between two words, X and Y,
by the mutual information estimate \cite{Church+Hanks:ACL89}:
\[\rm I(X,Y)\,=\,\log^+\;\frac{P(X\,|\,Y)}{P(X)}\;,\]
where $\rm P(X)$ is the occurrence density of word X in a whole corpus,
and the conditional probability $\rm P(X\,|\,Y)$ is the density of X
in a neighborhood of word Y.
Here the neighborhood is defined as 50 words before or after any
appearance of word Y.
(There is a variety of {\em neighborhood} definitions such as
``100 surrounding words'' \cite{Yarowsky:COLING92} and
``within a distance of no more than 3 words ignoring function words''
\cite{Dagan+al:ACL93}.)

The logarithm with `+' is defined to be 0 for an argument less than 1.
Negative estimates were neglected because they are mostly accidental
except when X and Y are frequent enough \cite{Church+Hanks:ACL89}.

A co-occurence vector of a word is defined as the list of co-occurrence
likelihood of the word with a certain set of origin words.
We used the same set of origin words as for the distance vectors.

\[\rm CV[w]\;=\;\left(\begin{array}{c}\rm I(w,O_1)\\\rm I(w,O_2)
\\\vdots\\\vdots\\\rm I(w,O_m)\end{array} \right)\hspace{5mm}\]

\centerline{\small\bf Co-occurrence Vector.}

When the frequency of X or Y is zero, we can not measure their
co-occurence likelihood, and such cases are not exceptional.
This sparseness problem is well-known and serious in the co-occurrence
statistics.
We used as a corpus the 1987 Wall Street Journal in the CD-ROM I
\shortcite{ACL-CD-ROM-1}, which has a total of 20M words.
The number of words which appeared at least once was about 50\%
of the total 62K head words of CED, and the percentage of
the {\it word-origin} pairs which appeared at least once was
about 16\% of total 62K $\times$ 1K (=62M) pairs.
When the co-occurrence likelihood can not be measured,
the value $\rm I(X,Y)$ was set to 0.

\section{Experimental Results}

We compared the two vector representations by using them for
the following two semantic tasks.
The first is word sense disambiguation (WSD) based on the similarity of
context vectors;
the second is the learning of \tmppos\ or \tmpneg\ meanings
from example words.

With WSD, the precision by using co-occurrence vectors
from a 20M words corpus was higher than by using distance vectors
from the CED.

\subsection{Word Sense Disambiguation}

Word sense disambiguation is a serious semantic problem.
A variety of approaches have been proposed for solving it.
For example, V\'{e}ronis and Ide \shortcite{Veronis+Ide:COLING90}
used reference networks as neural networks,
Hearst \shortcite{Hearst:NHD91} used (shallow) syntactic similarity
between contexts,
Cowie {\it et al.} \shortcite{Cowie+al:COLING92} used simulated
annealing for quick parallel disambiguation, and
Yarowsky \shortcite{Yarowsky:COLING92} used co-occurrence statistics
between words and thesaurus categories.

Our disambiguation method is based on the similarity of context vectors,
which was originated by Wilks {\it et al.} \shortcite{Wilks+al:MT90}.
In this method, a context vector is the sum of its constituent word
vectors (except the target word itself).
That is, the context vector for context,
\[\rm C:\;...\;w_{-N}\;...\;w_{-1}\;w\;w_{1}\;...\;w_{N'}\;...\;,\]
is
\[\rm V(C)\;=\;\sum_{i\,=\,-N}^{N'}\;V(w_{i})\, .\]
The similarity of contexts is measured by the angle of their vectors
(or actually the inner product of their normalized vectors).
\[\rm sim(C_1, C_2)\;=\;
        \frac{V(C_1)}{|V(C_1)|}\,\cdot\,\frac{V(C_2)}{|V(C_2)|}\;.\]

Let word \(\rm w\) have senses \(\rm s_1, s_2, ..., s_m\), and
each sense have the following context examples.
\vskip3mm
\centerline{
  \begin{tabular}{cl}
    Sense&Context Examples\\[2mm]\hline
    \(\rm s_1\)&\(\rm C_{1\,1},\;C_{1\,2},\;...\;C_{1\,n_1}\)\\
    \(\rm s_2\)&\(\rm C_{2\,1},\;C_{2\,2},\;...\;C_{2\,n_2}\)\\
    \(\vdots\) &\hspace{10mm}\(\vdots\)\\
    \(\rm s_m\)&\(\rm C_{m\,1},\;C_{m\,2},\;...\;C_{m\,n_m}\)\\
  \end{tabular}}
\vskip3mm

We infer that the sense of word \(\rm w\) in an arbitrary context \(\rm C\)
is \(\rm s_i\) if for some j the similarity, \(\rm sim(C,C_{i\,j})\),
is maximum among all the context examples.

Another possible way to infer the sense is to choose sense \(\rm s_i\)
such that the average of \(\rm sim(C,C_{i\,j})\) over
\(\rm j\,=\,1,2,...,n_i\) is maximum.
We selected the first method because a peculiarly similar example is more
important than the average similarity.

\newfignmb{\figwsdws}
Figure {\the\figwsdws} (next page) shows the disambiguation precision
for 9 words.
For each word, we selected two senses shown over each graph.
These senses were chosen because they are clearly different and
we could collect sufficient number (more than 20) of context examples.
The names of senses were chosen from the category names in
Roget's International Thesaurus, except {\it organ}'s.

The results using distance vectors are shown by dots
(\tmpd\hspace{1mm}\tmpd\hspace{1mm}\tmpd),
and using co-occurrence vectors from the 1987 WSJ (20M words)
by circles (\tmpca\hspace{1mm}\tmpca\hspace{1mm}\tmpca).

A context size (x-axis) of, for example, 10 means 10 words before the
target word and 10 words after the target word.
We used 20 examples per sense; they were taken from the 1988 WSJ.
The test contexts were from the 1987 WSJ:
The number of test contexts varies from word to word (100 to 1000).
The precision is the simple average of the respective precisions for
the two senses.

The results of Fig.\,{\the\figwsdws} show that the precision by
using co-occurrence vectors are higher than that by using distance
vectors except two cases, {\it interest} and {\it customs}.
And we have not yet found a case where the distance vectors give higher
precision. Therefore we conclude that co-occurrence vectors are
advantageous over distance vectors to WSD based on the context similarity.

The sparseness problem for co-occurrence vectors is not serious in this
case because each context consists of plural words.

\subsection{Learning of {\it \tmppos-or-\tmpneg}}

\newfignmb{\figgb}
\newtablenmb{\tablegbex}
\newtablenmb{\tablegbtest}
Another experiment using the same two vector representations
was done to measure the learning of \tmppos\ or \tmpneg\ meanings.
Figure {\the\figgb} shows the changes in the precision
(the percentage of agreement with the authors' combined judgement).
The x-axis indicates the number of example words for each
\tmppos\ or \tmpneg\ pair.
Judgement was again done by using the nearest example.
The example and test words are shown in Tables {\the\tablegbex} and
{\the\tablegbtest}, respectively.

\begin{figure}
\vskip5mm
\centerline{\unitlength=0.55mm
\begin{picture}(120,112)(0,-12)
\put(0,0){\line(1,0){120}}
\put(0,0){\line(0,1){100}}
\multiput(4,0)(4,0){30}{\line(0,1){1.5}}
\multiput(40,0)(40,0){3}{\line(0,2){2.0}}
\put(38,-6){10}\put(78,-6){20}\put(118,-6){30}
\put(70,-14){\makebox(0,0)[b]{\small\bf number of example pairs}}
\multiput(0,10)(0,10){10}{\line(1,0){2}}
\put(-12,48){50\%}\put(-15,98){100\%}
\put(4,72){\tmpd}
\put(8,60){\tmpd}
\put(12,66){\tmpd}
\put(16,65){\tmpd}
\put(20,67){\tmpd}
\put(24,65){\tmpd}
\put(28,72){\tmpd}
\put(32,72){\tmpd}
\put(36,71){\tmpd}
\put(40,71){\tmpd}
\put(44,72){\tmpd}
\put(48,79){\tmpd}
\put(52,75){\tmpd}
\put(56,70){\tmpd}
\put(60,75){\tmpd}
\put(64,75){\tmpd}
\put(68,75){\tmpd}
\put(72,75){\tmpd}
\put(76,75){\tmpd}
\put(80,80){\tmpd}
\put(84,80){\tmpd}
\put(88,80){\tmpd}
\put(92,80){\tmpd}
\put(96,80){\tmpd}
\put(100,80){\tmpd}
\put(104,80){\tmpd}
\put(108,80){\tmpd}
\put(112,81){\tmpd}
\put(116,81){\tmpd}
\put(120,81){\tmpd}
\put(4,49){\tmpca}
\put(8,63){\tmpca}
\put(12,55){\tmpca}
\put(16,55){\tmpca}
\put(20,55){\tmpca}
\put(24,55){\tmpca}
\put(28,52){\tmpca}
\put(32,54){\tmpca}
\put(36,51){\tmpca}
\put(40,51){\tmpca}
\put(44,50){\tmpca}
\put(48,53){\tmpca}
\put(52,57){\tmpca}
\put(56,57){\tmpca}
\put(60,57){\tmpca}
\put(64,57){\tmpca}
\put(68,57){\tmpca}
\put(72,57){\tmpca}
\put(76,57){\tmpca}
\put(80,54){\tmpca}
\put(84,54){\tmpca}
\put(88,54){\tmpca}
\put(92,54){\tmpca}
\put(96,54){\tmpca}
\put(100,54){\tmpca}
\put(104,54){\tmpca}
\put(108,49){\tmpca}
\put(112,50){\tmpca}
\put(116,50){\tmpca}
\put(120,51){\tmpca}
\multiput(0,50)(4,0){30}{\line(1,0){2.0}}
\multiput(0,80)(4,0){30}{\line(1,0){2.0}}
\put(50,10){
  \makebox(0,0)[b]{\small\bf
   \begin{tabular}[b]{c@{\hspace{3mm}}l}
     \tmpca & co-oc. vector (20M)\\
     \tmpd  & distance vector
   \end{tabular}}}
\end{picture}
}
\vskip2mm
\centerline{{\bf Fig. \the\figgb}\hspace{2mm}
  \vtop{\hbox{Learning of {\it \tmppos-or-\tmpneg}.}}}
\vskip3mm
\end{figure}

In this case, the distance vectors were advantageous.
The precision by using distance vectors increased to about 80\% and then
leveled off,
while the precision by using co-occurrence vectors stayed around 60\%.
We can therefore conclude that the property of {\it \tmppos-or-\tmpneg} is
reflected in distance vectors more strongly than in co-occurrence vectors.
The sparseness problem is supposed to be a major factor in this case.

\vskip3mm
\vtop{
\centerline{{\small\bf Table \the\tablegbex}\hspace{3mm} Example pairs.}
\vskip2mm
\centerline{\footnotesize
\hbox{
\begin{tabular}{rll}
   &{\normalsize\tmppos}&{\normalsize\tmpneg}\\
 1 & true  &  false\\
 2 & new  &  wrong\\
 3 & better  &  disease\\
 4 & clear  &  angry\\
 5 & pleasure  &  noise\\
 6 & correct  &  pain\\
 7 & pleasant  &  lose\\
 8 & suitable  &  destroy\\
 9 & clean  &  dangerous\\
10 & advantage\hspace{-3mm} &  harm\\
11 & love  &  kill\\
12 & best  &  fear\\
13 & successful\hspace{-3mm} &  war\\
14 & attractive\hspace{-3mm} &  ill\\
15 & powerful  &  foolish\\
\end{tabular}
\hspace{-5mm}
\begin{tabular}{rll}
   &{\normalsize\tmppos}&{\normalsize\tmpneg}\\
16 & properly  &  crime\\
17 & succeed  &  die\\
18 & worth  &  violent\\
19 & friendly  &  hurt\\
20 & useful  &  punishment\\
21 & success  &  poor\\
22 & interesting  &  badly\\
23 & active  &  fail\\
24 & polite  &  suffering\\
25 & win  &  enemy\\
26 & improve  &  rude\\
27 & favour  &  danger\\
28 & development\hspace{-3mm} &  anger\\
29 & happy  &  waste\\
30 & praise  &  doubt\\
\end{tabular}
}}}

\vskip3mm
\vtop{
\centerline{{\small\bf Table \the\tablegbtest}\hspace{3mm} Test words.}
\vskip3mm
\vbox{{\xcomment{\normalsize}{} \tmppos}\hspace{5mm}(20 words)\vspace{1.5mm}
\xcomment{\footnotesize}{}\def\tmp{\hspace{\xcomment{0.7mm}{2mm}}}
\halign{\hspace{3mm}#&\tmp#&\tmp#&\tmp#&\tmp#\cr
balanced & elaborate & elation & eligible & enjoy\cr
fluent & honorary & honourable & hopeful & hopefully\cr
influential & interested & legible & lustre & normal\cr
recreation & replete & resilient & restorative & sincere\cr}}
\vskip3mm
\vbox{{\xcomment{\normalsize}{} \tmpneg}\hspace{5mm}(30 words)\vspace{1.5mm}
\xcomment{\footnotesize}{}\def\tmp{\hspace{\xcomment{0.7mm}{2mm}}}
\halign{\hspace{3mm}#&\tmp#&\tmp#&\tmp#&\tmp#\cr
confusion & cuckold & dally & damnation & dull\cr
ferocious & flaw & hesitate & hostage & huddle\cr
inattentive & liverish & lowly & mock & neglect\cr
queer & rape & ridiculous & savage & scanty\cr
sceptical & schizophrenia & scoff & scruffy & shipwreck\cr
superstition & sycophant & trouble & wicked & worthless\cr}}
}

\addtocounter{page}{1}

\subsection{Supplementary Data}

In the experiments discussed above, the corpus size for co-occurrence
vectors was set to 20M words ('87 WSJ)
and the vector dimension for both co-occurrence and distance vectors was
set to 1000.
Here we show some supplementary data that support these parameter
settings.

{\bf a. Corpus size (for co-occurrence vectors)}

\newfignmb{\figcs}
\def\tmpcsword{issue}
\def\tmpcsws{10}
\def\tmpcsnea{10}
Figure {\the\figcs} shows the change in disambiguation precision
as the corpus size for co-occurrence statistics increases from 200
words to 20M words.
(The words are {\it suit}, {\it issue} and {\it race},
the context size is \tmpcsws,
and the number of examples per sense is \tmpcsnea.)
These three graphs level off after around 1M words.
Therefore, a corpus size of 20M words is not too small.

\vskip5mm
\centerline{
 \xcomment{
  \graphcs{
   {\def\tmpnea{$\circ$}
\put(6,48){\tmpnea}
\put(14,50){\tmpnea}
\put(20,63){\tmpnea}
\put(26,64){\tmpnea}
\put(34,62){\tmpnea}
\put(40,69){\tmpnea}
\put(46,59){\tmpnea}
\put(54,71){\tmpnea}
\put(60,77){\tmpnea}
\put(66,84){\tmpnea}
\put(74,81){\tmpnea}
\put(80,86){\tmpnea}
\put(86,85){\tmpnea}
\put(94,83){\tmpnea}
\put(100,83){\tmpnea}
\put(106,84){\tmpnea}
    }
   {\def\tmpnea{$\diamond$}
\put(6,49){\tmpnea}
\put(14,50){\tmpnea}
\put(20,46){\tmpnea}
\put(26,58){\tmpnea}
\put(34,49){\tmpnea}
\put(40,58){\tmpnea}
\put(46,55){\tmpnea}
\put(54,53){\tmpnea}
\put(60,59){\tmpnea}
\put(66,60){\tmpnea}
\put(74,66){\tmpnea}
\put(80,71){\tmpnea}
\put(86,76){\tmpnea}
\put(94,79){\tmpnea}
\put(100,80){\tmpnea}
\put(106,77){\tmpnea}
   }
   {\def\tmpnea{$\ast$}
\put(6,55){\tmpnea}
\put(14,50){\tmpnea}
\put(20,58){\tmpnea}
\put(26,71){\tmpnea}
\put(34,78){\tmpnea}
\put(40,76){\tmpnea}
\put(46,78){\tmpnea}
\put(54,80){\tmpnea}
\put(60,93){\tmpnea}
\put(66,94){\tmpnea}
\put(74,90){\tmpnea}
\put(80,96){\tmpnea}
\put(86,95){\tmpnea}
\put(94,94){\tmpnea}
\put(100,94){\tmpnea}
\put(106,94){\tmpnea}
   }
   \put(50,10){
    \makebox(0,0)[b]{\small\bf
     \begin{tabular}[b]{c@{\hspace{3mm}}l}
       $\ast$     & suit\\
       $\circ$   & issue\\
       $\diamond$ & race\\
     \end{tabular}}}
   \put(100,-10){\makebox(0,0)[t]{\small\bf (word)}}
   \put(60,-12){\makebox(0,0)[t]{\small\bf corpus size}}
  }}{\bf (Fig. \the\figcs)}}
\vskip\xcomment{7mm}{3mm}
\centerline{{\bf Fig. \the\figcs}\hspace{2mm}
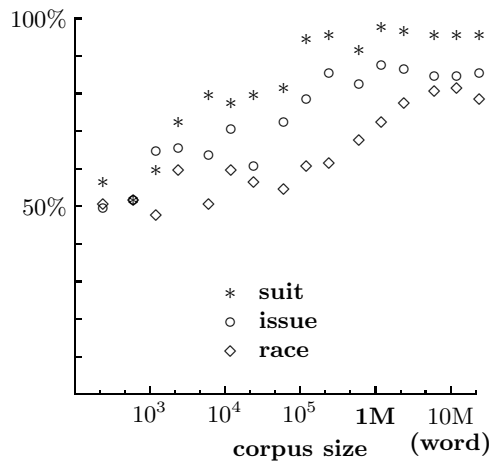
  \vtop{\hbox{Dependence of the disambiguation precision}
        \hbox{on the corpus size for co-occurrence vectors.}
        \hbox{\hspace{5mm}context size: \tmpcsws,}
        \hbox{\hspace{5mm}number of examples: \tmpcsnea/sense,}
        \hbox{\hspace{5mm}vector dimension: 1000.}}}
\vskip3mm

{\bf b. Vector Dimension}

\newfignmb{\figvd}\def\tmpvdws{10}\def\tmpvdne{10}

Figure {\the\figvd} (next page) shows the dependence of disambiguation
precision on the vector dimension for (i) co-occurrence and
(ii) distance vectors.
As for co-occurrence vectors, the precision levels off near a dimension
of 100. Therefore, a dimension size of 1000 is sufficient or even
redundant.
However, in the distance vector's case, it is not clear whether
the precision is leveling or still increasing around 1000 dimension.

\section{Conclusion}

\begin{itemize}
\item
A comparison was made of co-occurrence vectors from large text corpora
and of distance vectors from dictionary definitions.
\item
For the word sense disambiguation based on the context similarity,
co-occurrence vectors from the 1987 Wall Street Journal (20M total words)
was advantageous over distance vectors from the Collins English Dictionary
({\cedsizehead} head words + {\cedsizedef} definition words).
\item
For learning \tmppos\ or \tmpneg\ meanings from example words,
distance vectors gave remarkably higher precision than co-occurrence
vectors.
This suggests, though further investigation is required, that distance
vectors contain some different semantic information from
co-occurrence vectors.
\end{itemize}

\centerline{\small\bf (i) {\rm by} co-oc. vectors}
\vskip3mm
\centerline{
 \graphvd{
   {\def\tmpca{$\circ$}
\put(10,68){\tmpca}
\put(20,77){\tmpca}
\put(30,77){\tmpca}
\put(40,79){\tmpca}
\put(50,86){\tmpca}
\put(60,88){\tmpca}
\put(70,86){\tmpca}
\put(80,83){\tmpca}
\put(90,83){\tmpca}
\put(100,84){\tmpca}
   }
   {\def\tmpca{$\diamond$}
\put(10,51){\tmpca}
\put(20,59){\tmpca}
\put(30,59){\tmpca}
\put(40,58){\tmpca}
\put(50,70){\tmpca}
\put(60,75){\tmpca}
\put(70,72){\tmpca}
\put(80,74){\tmpca}
\put(90,75){\tmpca}
\put(100,77){\tmpca}
   }
   {\def\tmpca{$\ast$}
\put(10,54){\tmpca}
\put(20,91){\tmpca}
\put(30,87){\tmpca}
\put(40,91){\tmpca}
\put(50,93){\tmpca}
\put(60,96){\tmpca}
\put(70,95){\tmpca}
\put(80,95){\tmpca}
\put(90,94){\tmpca}
\put(100,94){\tmpca}
   }
   \put(50,10){
    \makebox(0,0)[b]{\small\bf
     \begin{tabular}[b]{c@{\hspace{3mm}}l}
       $\ast$     & suit\\
       $\circ$   & issue\\
       $\diamond$ & race\\
     \end{tabular}}}
   \put(60,-10){\makebox(0,0)[t]{\small\bf vector dimension}}}}
\vskip8mm
\centerline{\small\bf (ii) {\rm by} distance vectors}
\centerline{
 \graphvd{
   \def\tmpca{}
   {\def\tmpd{$\circ$}
\put(10,50){\tmpd}
\put(20,42){\tmpd}
\put(30,61){\tmpd}
\put(40,58){\tmpd}
\put(50,60){\tmpd}
\put(60,65){\tmpd}
\put(70,69){\tmpd}
\put(80,69){\tmpd}
\put(90,67){\tmpd}
\put(100,67){\tmpd}
   }
   {\def\tmpd{$\diamond$}
\put(10,48){\tmpd}
\put(20,57){\tmpd}
\put(30,50){\tmpd}
\put(40,50){\tmpd}
\put(50,53){\tmpd}
\put(60,62){\tmpd}
\put(70,65){\tmpd}
\put(80,72){\tmpd}
\put(90,72){\tmpd}
\put(100,75){\tmpd}
   }
   {\def\tmpd{$\ast$}
\put(10,63){\tmpd}
\put(20,60){\tmpd}
\put(30,61){\tmpd}
\put(40,72){\tmpd}
\put(50,75){\tmpd}
\put(60,78){\tmpd}
\put(70,81){\tmpd}
\put(80,85){\tmpd}
\put(90,89){\tmpd}
\put(100,86){\tmpd}
   }
   \put(50,10){
    \makebox(0,0)[b]{\small\bf
     \begin{tabular}[b]{c@{\hspace{3mm}}l}
       $\ast$     & suit\\
       $\circ$   & issue\\
       $\diamond$ & race\\
     \end{tabular}}}
   \put(60,-10){\makebox(0,0)[t]{\small\bf vector dimension}}}}
\vskip7mm
\centerline{{\bf Fig. \the\figvd}\hspace{2mm}
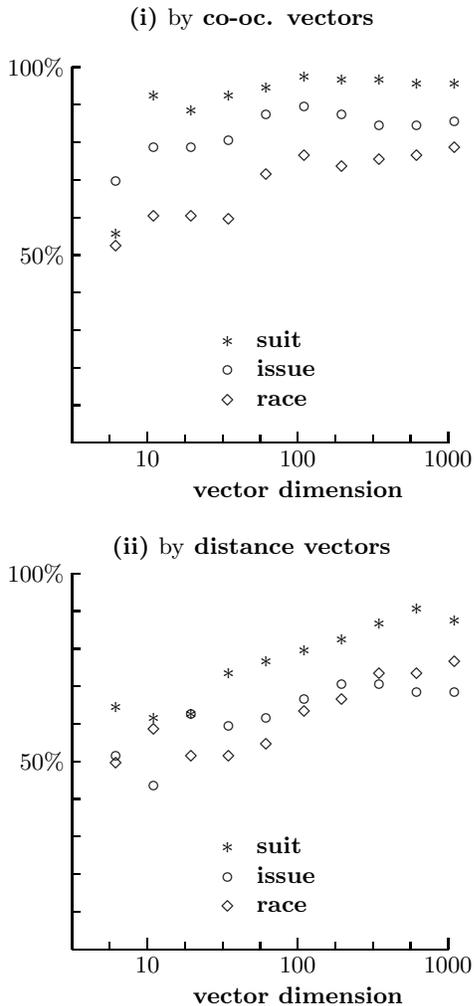
  \parbox[t]{70mm}{Dependence on vector dimension for
(i) co-occurrence vectors and (ii) distance vectors.\\
context size: \tmpvdws,\hspace{2mm} examples: \tmpvdne/sense,\\
corpus size for co-oc. vectors: 20M word.}}

\Comment{
\bibliographystyle{\bibdir/acl94}
\bibliography{\bibdir/niwa}
}

\vfil\break
\setcounter{page}{307}
\def\Fmesh#1{}
\def\Fgraph#1{#1}
\onecolumn

\vbox to \textheight{
\def\tmpcb{}
\unitlength=0.4mm
\def\graphA#1{
  \begin{picture}(110,100)(0,0)
    \put(0,0){\line(1,0){110}}
    \put(0,0){\line(0,1){100}}
    \multiput(0,10)(0,10){10}{\line(1,0){2}}
    \def\tmpym##1##2{\put(-2.5,##1){\makebox(0,0)[r]{\small ##2}}}
    \tmpym{50}{50}\tmpym{100}{100}
    \multiput(5,0)(5,0){10}{\line(0,1){2}}
    \multiput(65,0)(15,0){4}{\line(0,1){2}}
    \def\tmpxm##1##2{\put(##1,-2.5){\makebox(0,0)[t]{\small ##2}}}
    \tmpxm{25}{5}\tmpxm{50}{10}\tmpxm{65}{20}
    \tmpxm{80}{30}\tmpxm{95}{40}\tmpxm{110}{50}
    #1
  \end{picture}}
\def\graphE#1#2#3#4#5{
{\unitlength=1mm
  \begin{picture}(44,52)(0,0)
    \put(0,0){\makebox(0,52)[lt]{\bf #1\ \ {\scriptsize\bf (#2 / #3)}}}
    \put(0,0){\unitlength=0.4mm\graphA{#4}}
    #5
  \end{picture}}}
\def\tmpexp{
 \put(22,5){
  \makebox(0,0)[b]{\small\bf
   \begin{tabular}[b]{c@{\hspace{3mm}}l}
     \tmpca & co-oc. vector  \\
     \tmpd  & distance vector
   \end{tabular}}}
 \put(25,-7){\makebox(0,0)[b]{\normalsize\bf context size}}
 \put(-2,44){\makebox(0,0)[r]{\normalsize\bf \%}}}
\Fmesh{\hrule}
\def\neps{10}
\vfil
\unitlength=1mm
\centerline{
  \begin{picture}(160,60)(0,-8)
    \Fmesh{\multiput(0,-8)(160,0){2}{\line(0,1){58}}}
    \put(  0,0){\graphE{suit}{CLOTHING}{LAWSUIT}{
\put(5,83){\tmpd}\put(10,86){\tmpd}\put(15,89){\tmpd}\put(20,85){\tmpd}
\put(25,85){\tmpd}\put(30,91){\tmpd}\put(35,91){\tmpd}\put(40,88){\tmpd}
\put(45,82){\tmpd}\put(50,86){\tmpd}\put(65,84){\tmpd}\put(80,78){\tmpd}
\put(95,79){\tmpd}\put(110,86){\tmpd}
\put(5,90){\tmpca}\put(10,84){\tmpca}\put(15,91){\tmpca}\put(20,91){\tmpca}
\put(25,89){\tmpca}\put(30,92){\tmpca}\put(35,94){\tmpca}\put(40,97){\tmpca}
\put(45,94){\tmpca}\put(50,94){\tmpca}\put(65,94){\tmpca}\put(80,95){\tmpca}
\put(95,96){\tmpca}\put(110,96){\tmpca}
}{\tmpexp}}
    \put( 55,0){\graphE{organ}{BODY}{MUSIC}{
\put(5,60){\tmpd}\put(10,65){\tmpd}\put(15,66){\tmpd}\put(20,65){\tmpd}
\put(25,75){\tmpd}\put(30,66){\tmpd}\put(35,64){\tmpd}\put(40,61){\tmpd}
\put(45,56){\tmpd}\put(50,72){\tmpd}\put(65,74){\tmpd}\put(80,86){\tmpd}
\put(95,89){\tmpd}\put(110,94){\tmpd}
\put(5,60){\tmpca}\put(10,72){\tmpca}\put(15,83){\tmpca}\put(20,91){\tmpca}
\put(25,91){\tmpca}\put(30,89){\tmpca}\put(35,89){\tmpca}\put(40,89){\tmpca}
\put(45,88){\tmpca}\put(50,91){\tmpca}\put(65,92){\tmpca}\put(80,91){\tmpca}
\put(95,92){\tmpca}\put(110,90){\tmpca}
}{}}
    \put(110,0){\graphE{issue}{EMERGENCE}{TOPIC}{
\put(5,65){\tmpd}\put(10,68){\tmpd}\put(15,67){\tmpd}\put(20,66){\tmpd}
\put(25,64){\tmpd}\put(30,65){\tmpd}\put(35,68){\tmpd}\put(40,69){\tmpd}
\put(45,72){\tmpd}\put(50,67){\tmpd}\put(65,78){\tmpd}\put(80,80){\tmpd}
\put(95,74){\tmpd}\put(110,76){\tmpd}
\put(5,60){\tmpca}\put(10,74){\tmpca}\put(15,76){\tmpca}\put(20,83){\tmpca}
\put(25,81){\tmpca}\put(30,80){\tmpca}\put(35,84){\tmpca}\put(40,85){\tmpca}
\put(45,85){\tmpca}\put(50,84){\tmpca}\put(65,81){\tmpca}\put(80,77){\tmpca}
\put(95,78){\tmpca}\put(110,80){\tmpca}
}{}}
  \end{picture}}
\vfil
\centerline{\begin{picture}(160,60)(0,-8)
  \put(  0,0){\graphE{tank}{CONTAINER}{VEHICLE}{
\put(5,72){\tmpd}\put(10,57){\tmpd}\put(15,59){\tmpd}\put(20,70){\tmpd}
\put(25,73){\tmpd}\put(30,68){\tmpd}\put(35,68){\tmpd}\put(40,74){\tmpd}
\put(45,72){\tmpd}\put(50,72){\tmpd}\put(65,73){\tmpd}\put(80,75){\tmpd}
\put(95,75){\tmpd}\put(110,70){\tmpd}
\put(5,70){\tmpca}\put(10,74){\tmpca}\put(15,79){\tmpca}\put(20,81){\tmpca}
\put(25,80){\tmpca}\put(30,83){\tmpca}\put(35,82){\tmpca}\put(40,77){\tmpca}
\put(45,89){\tmpca}\put(50,85){\tmpca}\put(65,85){\tmpca}\put(80,80){\tmpca}
\put(95,78){\tmpca}\put(110,75){\tmpca}
}{\tmpexp}}
  \put( 55,0){\graphE{order}{COMMAND}{DEMAND}{
\put(5,69){\tmpd}\put(10,67){\tmpd}\put(15,64){\tmpd}\put(20,57){\tmpd}
\put(25,65){\tmpd}\put(30,72){\tmpd}\put(35,71){\tmpd}\put(40,61){\tmpd}
\put(45,64){\tmpd}\put(50,68){\tmpd}\put(65,60){\tmpd}\put(80,55){\tmpd}
\put(95,60){\tmpd}\put(110,64){\tmpd}
\put(5,65){\tmpca}\put(10,61){\tmpca}\put(15,70){\tmpca}\put(20,72){\tmpca}
\put(25,66){\tmpca}\put(30,76){\tmpca}\put(35,74){\tmpca}\put(40,69){\tmpca}
\put(45,72){\tmpca}\put(50,72){\tmpca}\put(65,75){\tmpca}\put(80,71){\tmpca}
\put(95,79){\tmpca}\put(110,78){\tmpca}
}{}}
  \put(110,0){\graphE{address}{HABITAT}{SPEECH}{
\put(5,80){\tmpd}\put(10,67){\tmpd}\put(15,66){\tmpd}\put(20,73){\tmpd}
\put(25,62){\tmpd}\put(30,73){\tmpd}\put(35,75){\tmpd}\put(40,74){\tmpd}
\put(45,70){\tmpd}\put(50,68){\tmpd}\put(65,67){\tmpd}\put(80,71){\tmpd}
\put(95,66){\tmpd}\put(110,80){\tmpd}
\put(5,74){\tmpca}\put(10,73){\tmpca}\put(15,73){\tmpca}\put(20,70){\tmpca}
\put(25,71){\tmpca}\put(30,72){\tmpca}\put(35,73){\tmpca}\put(40,78){\tmpca}
\put(45,71){\tmpca}\put(50,67){\tmpca}\put(65,80){\tmpca}\put(80,83){\tmpca}
\put(95,79){\tmpca}\put(110,81){\tmpca}
}{}}
\end{picture}}
\vfil
\centerline{\begin{picture}(160,60)(0,-8)
  \put(  0,0){\graphE{race}{CLASS}{OPPOSITION}{
\put(5,68){\tmpd}\put(10,68){\tmpd}\put(15,67){\tmpd}\put(20,67){\tmpd}
\put(25,66){\tmpd}\put(30,69){\tmpd}\put(35,78){\tmpd}\put(40,74){\tmpd}
\put(45,71){\tmpd}\put(50,75){\tmpd}\put(65,70){\tmpd}\put(80,67){\tmpd}
\put(95,65){\tmpd}\put(110,62){\tmpd}
\put(5,61){\tmpca}\put(10,72){\tmpca}\put(15,69){\tmpca}\put(20,73){\tmpca}
\put(25,80){\tmpca}\put(30,74){\tmpca}\put(35,80){\tmpca}\put(40,83){\tmpca}
\put(45,80){\tmpca}\put(50,77){\tmpca}\put(65,81){\tmpca}\put(80,76){\tmpca}
\put(95,76){\tmpca}\put(110,73){\tmpca}
}{\tmpexp}}
  \put( 55,0){\graphE{customs}{HABIT(pl.)}{SERVICE}{
\put(5,73){\tmpd}\put(10,64){\tmpd}\put(15,70){\tmpd}\put(20,58){\tmpd}
\put(25,69){\tmpd}\put(30,71){\tmpd}\put(35,64){\tmpd}\put(40,58){\tmpd}
\put(45,65){\tmpd}\put(50,72){\tmpd}\put(65,69){\tmpd}\put(80,62){\tmpd}
\put(95,61){\tmpd}\put(110,66){\tmpd}
\put(5,68){\tmpca}\put(10,71){\tmpca}\put(15,82){\tmpca}\put(20,70){\tmpca}
\put(25,78){\tmpca}\put(30,77){\tmpca}\put(35,71){\tmpca}\put(40,68){\tmpca}
\put(45,73){\tmpca}\put(50,72){\tmpca}\put(65,60){\tmpca}\put(80,69){\tmpca}
\put(95,64){\tmpca}\put(110,67){\tmpca}
}{}}
  \put(110,0){\graphE{interest}{CURIOSITY}{DEBT}{
\put(5,80){\tmpd}\put(10,67){\tmpd}\put(15,72){\tmpd}\put(20,80){\tmpd}
\put(25,73){\tmpd}\put(30,72){\tmpd}\put(35,68){\tmpd}\put(40,72){\tmpd}
\put(45,69){\tmpd}\put(50,73){\tmpd}\put(65,66){\tmpd}\put(80,71){\tmpd}
\put(95,69){\tmpd}\put(110,71){\tmpd}
\put(5,82){\tmpca}\put(10,76){\tmpca}\put(15,72){\tmpca}\put(20,69){\tmpca}
\put(25,68){\tmpca}\put(30,68){\tmpca}\put(35,62){\tmpca}\put(40,76){\tmpca}
\put(45,71){\tmpca}\put(50,75){\tmpca}\put(65,66){\tmpca}\put(80,72){\tmpca}
\put(95,78){\tmpca}\put(110,72){\tmpca}
}{}}
\end{picture}}
\vskip 5mm\Fmesh{\hrule}\vfil
\centerline{\bf Fig. \the\figwsdws\hspace{3mm}
  \parbox[t]{125mm}{
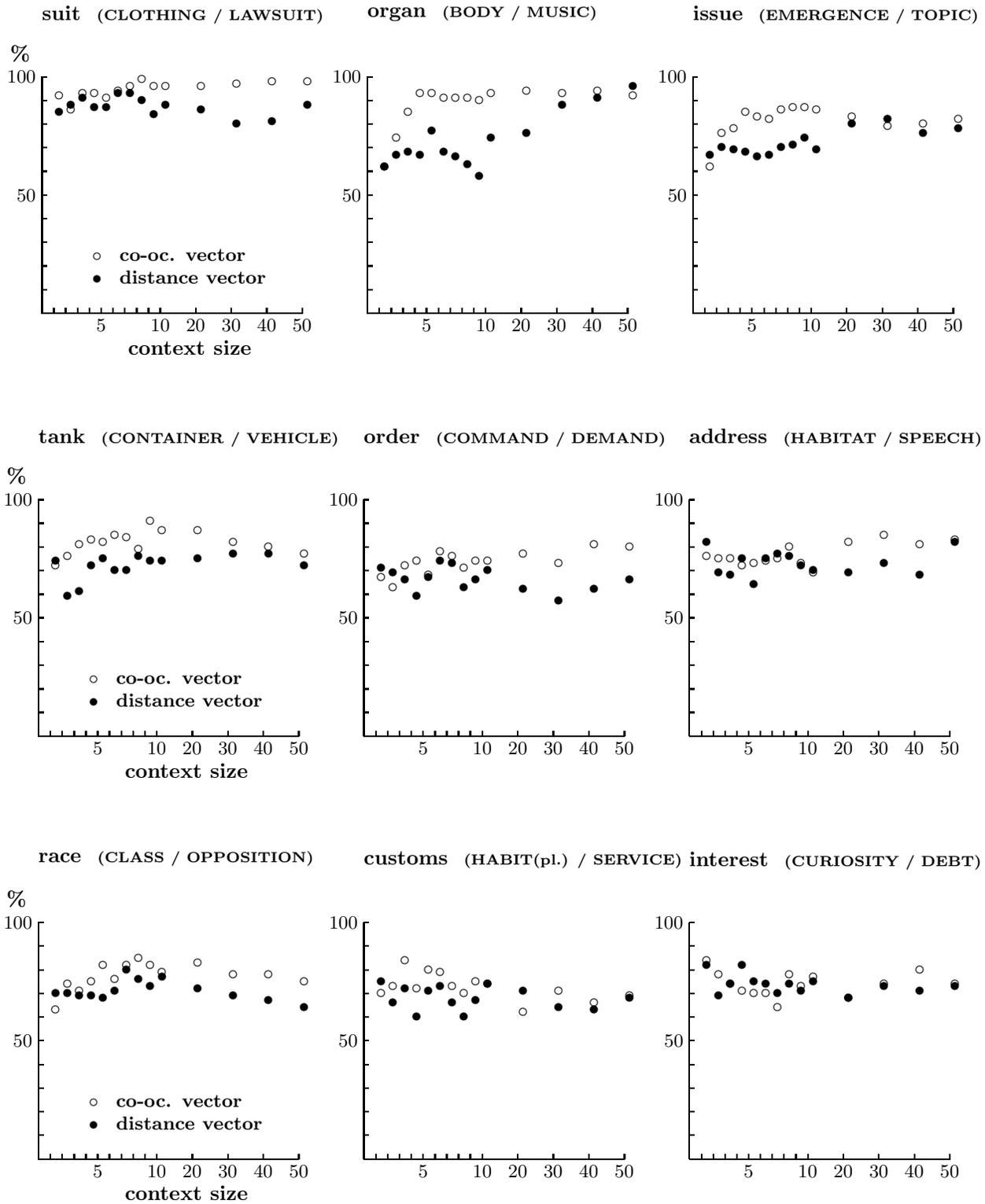
Disambiguation of 9 words by using co-occurrence vectors(\tmpca \tmpca \tmpca)
and by using distance vectors (\tmpd \tmpd \tmpd).
{\rm (The number of examples is {\neps} for each sense.)}}}
\vfil
\Fmesh{\hrule}
}
\end{document}